\documentclass[letterpaper, 10 pt, conference]{ieeeconf}  

\IEEEoverridecommandlockouts                              

\usepackage{graphicx}

\title{\LARGE \bf
Selective Unit-Cell Actuation in Lattice Structures for Distributed Morphology in Soft Robots}

\author{Trevor Exley$^{1}$, Altair Coutinho$^{1}$ and Lucia Beccai$^{1}$
\thanks{$^{1}$T. Exley, A. Coutinho, and L. Beccai are with Soft BioRobotics Perception Lab of the Istituto Italiano di Tecnologia (IIT), Genova 16163, Italy (lucia.beccai@iit.it)%
}%
}

\begin{document}

\maketitle
\thispagestyle{empty}
\pagestyle{empty}

\begin{abstract}

Soft lattice structures are increasingly used in robotics to tailor compliance and guide deformation; however, actuation is typically introduced at the device or module level, with actuators inserted into otherwise passive architectures. In this work, we move actuator-lattice co-design to the unit-cell scale. We present an embedded pneumatic unit cell that integrates curved-strut lattice geometry with a bidirectional bellow actuator within a single monolithic element. When tessellated, the lattice functions as a distributed actuation field in which global morphology is governed by spatial actuation patterns rather than uniform pressurization. Experimental characterization of $1\times1$, $2\times2$, and $3\times3$ tessellations demonstrates scalable displacement and force generation with repeatable cyclic performance. Selective actuation of unit cells in a $3\times3\times3$ array produces distinct global deformation modes, including bending and directional grasping, without altering hardware configuration. Additionally, coupling active and passive unit cells enables bending-driven crawling locomotion, demonstrating that heterogeneous tessellations can translate through asymmetric deformation. These results establish unit-cell-level actuation as a strategy for distributed morphing in lattice-based soft robots and provide a foundation for scalable, monolithic robotic architectures.
\end{abstract}

\section{INTRODUCTION}

\begin{figure*}[t]
    \centering
    \vspace{1em}
    \includegraphics[width=0.9\textwidth]{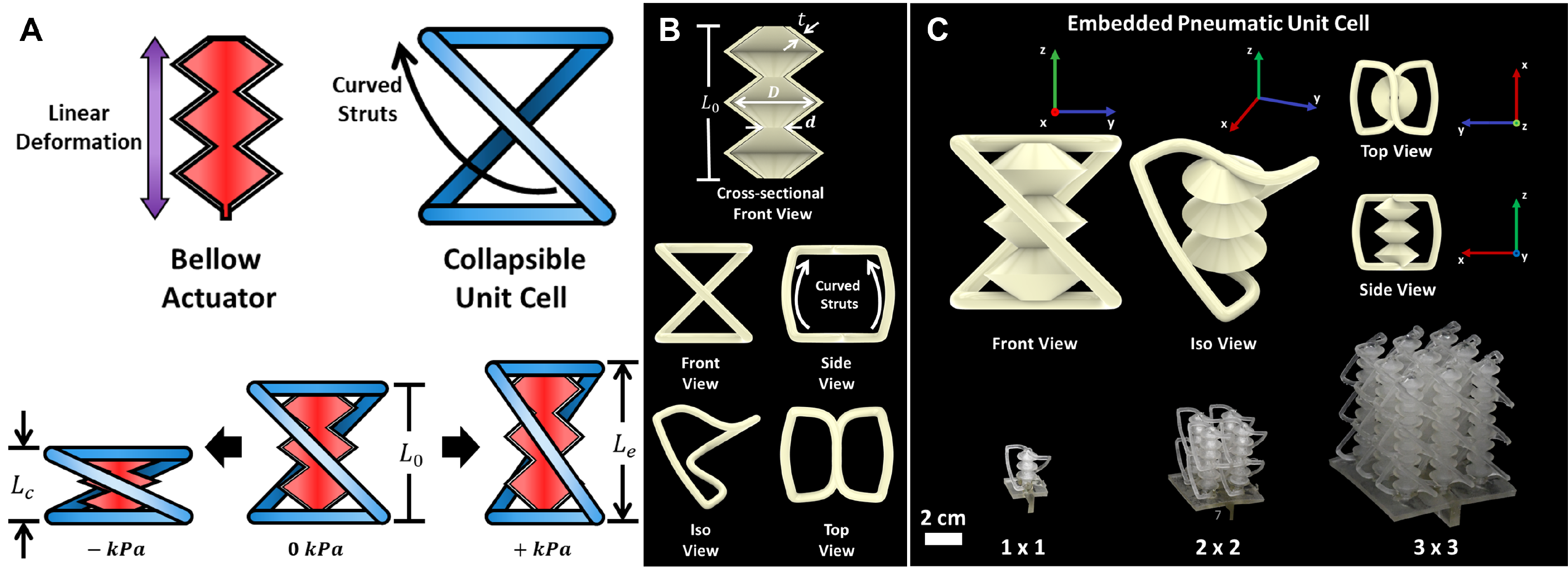}
    \caption{Design concept of the bellow actuator and collapsible unit cell in (A) 2D and (B) 3D with its main dimension parameters. (C) Embedded pneumatic unit cell shown in all views with printed prototypes of $1\times1$, $2\times2$, and $3\times3$ tesselations.}
    \label{fig:design}
\end{figure*}

Soft robots derive their advantages in safe, adaptive interaction from bodies that can continuously deform to conform around objects, navigate cluttered environments, and tolerate uncertainty \cite{schaffner3DPrintingRobotic2018,liSoftActuatorsRealworld2021}. Pneumatic actuation remains one of the most practical approaches for driving these systems due to its simplicity, high power-to-weight potential, and compatibility with elastomeric additive manufacturing \cite{zhangRoboticArtificialMuscles2019}. At the same time, achieving predictable and programmable deformation in pneumatic systems is fundamentally a coupled \emph{structure-actuation} design problem: internal pressure loads act normal to cavity surfaces, and without careful geometric and stiffness programming, deformation modes can be difficult to constrain or scale \cite{coutinhoMorphingSuctionCups2024}. Recent work on triply periodic channel architectures has demonstrated how carefully designed internal geometries enable scalable, single-material pneumatic linear actuators \cite{chenTriplyPeriodicChannels2022a}, while studies of periodically porous elastomers have revealed rich bifurcation and bistable behaviors under pneumatic loading \cite{liangBifurcationBistabilityPneumatically2023}. These results underscore the central role of geometry at the unit-cell level in governing global actuation behavior. Recent advances in reprogrammable soft pneumatic metamaterials further demonstrate how unit-cell asymmetry and modular architectural features can enable tunable multistability and multimodal deformation within tessellated structures \cite{rahmanReprogrammableSoftPneumatic2025}.

In parallel, architected lattices and mechanical metamaterials have emerged as powerful tools for tailoring stiffness, anisotropy, and deployability in soft systems. Fractal-inspired and hierarchical deployable lattices exhibit extreme shape change and tunable effective properties \cite{xiongFractalinspiredSoftDeployable2021}, while multiphase metamaterials enable negative expansion and programmable responses through structural design \cite{taoThreephaseSoftMechanical2024}. In soft robotics, lattice configuration has been leveraged to modulate joint stiffness \cite{wangStiffnessModulationSoft2021}, to realize reconfigurable feet for dexterous locomotion \cite{wangDexterousElectricaldrivenSoft2023a}, and to steer actuator bending via auxetic robotic skins \cite{puRoboticSkinsInspired2024a}. Beyond purely passive structures, electrothermally actuated lattice metamaterials \cite{zhangElectrothermallyActuatedLattice2025} and fabric-lattice artificial muscles that combine active air chambers with passive lattice layers \cite{yangHighStrokeHighOutputForceFabricLattice2024a} further demonstrate that cellular architectures can serve as substrates for embedding functionality.

Recent efforts within soft robotics have emphasized \emph{monolithic} fabrication strategies, in which structure and actuation are co-fabricated within a single printed body to reduce mechanical discontinuities and modeling uncertainty \cite{truninMELEGROSMonolithicElephantInspired2026, exleyMonolithicUnitsActuation2025}. These approaches highlight the benefits of embedding functionality directly into architected soft materials and establishing geometric co-design principles at the device level. However, in such systems, the lattice primarily interacts with the actuator as a strain-limiting layer. As a result, the actuator defines the dominant deformation mode, while the lattice informs or restricts that motion, rather than participating directly in distributed actuation at the unit-cell scale.

More broadly, most existing lattice-actuator integrations operate at the macro scale: an actuator (or set of actuators) is embedded within or attached to an otherwise passive lattice body, and the lattice geometry is tuned to guide the resulting global deformation. Modular approaches such as self-connecting soft blocks enable reconfigurable deformable lattice assemblies \cite{zhaoStarBlocksSoftActuated2023,parkReconfigurableShapeMorphing2022}, and pneumatically controlled lattices have recently been shown to exhibit tunable global mechanical behavior under internal pressurization \cite{zhuPneumaticallyControlledLattices2025}. Hybrid additive manufacturing strategies have enabled active lattice architectures using liquid crystal elastomers and other stimuli-responsive materials \cite{peng4DPrintingFreestanding2022a,jiang4DPrintingVat2026}. However, in most of these approaches, actuation and lattice topology are conceived as coupled but distinct layers of design: the lattice provides structure, and actuation is introduced into selected regions, skins, or modules.

In this work, we move actuator-lattice co-design down to the \emph{unit-cell level}. Rather than embedding discrete actuators within an otherwise passive lattice, we develop a pneumatic lattice architecture in which the actuator and structural geometry are co-designed within the unit cell itself. Each unit cell simultaneously serves as a load-bearing architectural element and an addressable actuation primitive. By tessellating these embedded pneumatic unit cells within a single monolithic elastomeric body, the lattice becomes an actively programmable structure. Global deformation is no longer dictated solely by uniform pressurization, but by the spatial distribution of pressure across the cellular array. Selective actuation patterns across the lattice therefore determine the resulting global morphology.

This unit-cell perspective extends previous concepts of pneumatic metamaterials \cite{chenTriplyPeriodicChannels2022a,liangBifurcationBistabilityPneumatically2023} by embedding actuation directly into the unit-cell, rather than repeating a single actuation topology or inserting actuators into a passive scaffold. It also builds on monolithic soft robotic design principles \cite{truninMELEGROSMonolithicElephantInspired2026,exleyMonolithicUnitsActuation2025} by unifying structure and actuation at the smallest repeatable scale of the lattice. By co-designing geometry and pneumatic function at the unit-cell level, we realize single-material morphing structures whose global behavior is governed not only by pressure magnitude but by actuation topology.

The contributions of this work are threefold: (1) a co-design framework that integrates pneumatic actuation directly into a curved-strut lattice unit cell; (2) a strategy for achieving programmable global morphologies through selective actuation of tessellated unit cells; and (3) experimental demonstrations showing that a single monolithic printed lattice can achieve distinct functional behaviors, including directional grasping through selective actuation and bending-driven locomotion through active-passive tessellation, without altering the underlying hardware architecture.

\section{METHODOLOGY}

\begin{figure*}[t]
    \centering
    \vspace{1em}
    \includegraphics[width=0.9\textwidth]{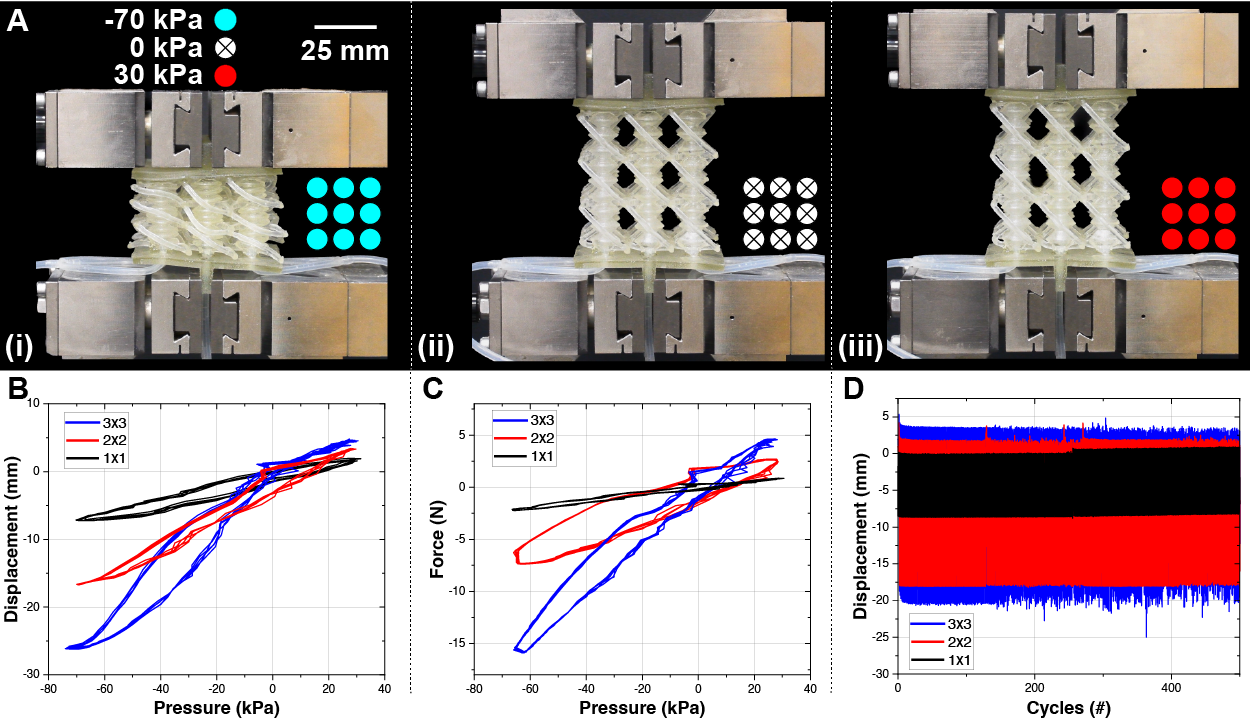}
    \caption{(A) Experimental setup and testing of the EPUC during (i) contraction, (ii) rest, and (iii) expansion in isotonic tests. Representative results for (B) isotonic displacement--pressure response, (C) isometric force--pressure response, and (D) free-moving cyclic displacement response over 500 cycles for $1\times1$, $2\times2$, and $3\times3$ specimens used to assess fatigue.}
    \label{fig:exp-setup}
\end{figure*}

\subsection{Unit Cell Co-Design}

The embedded pneumatic unit cell (EPUC) (Fig.~\ref{fig:design}) integrates lattice geometry and pneumatic actuation within a single continuous body. Rather than inserting an actuator into a pre-designed lattice, the actuator and struts are co-designed at the unit-cell level.

The pneumatic actuator within the EPUC is a bellow-type actuator selected for its ability to generate bidirectional linear deformation under both positive pressure and vacuum conditions. The bellow actuator in this work has an initial length (\(L_0\)) of 18~mm, an inner diameter (\(d\)) and an outer diameter (\(D\)) of 3~mm and 11~mm, respectively. The membrane thickness (\(t\)) of the actuator membrane is 0.5~mm as it provides sufficient compliance to enable axial expansion and contraction while preventing lateral buckling and limiting radial expansion under pressurization.

The collapsible unit cell consists of two opposing diagonal struts that bow outward from the centerline. Each cell has an edge length of 18~mm and strut diameter of 2~mm. Curved struts were selected based on prior studies showing that initial curvature enables controlled buckling and guided deformation in lattice members \cite{zhuLatticeMaterialsComposed2018,tuminoMechanicalPropertiesBCC2024,sasagawaFlexibleLatticeStructure2025}. In this design, the curvature promotes symmetric lateral expansion during axial contraction.

When stacked vertically, the outward-bowing diagonal struts rotate about their base during axial contraction, converting vertical shortening into transverse expansion. This scissor-like interaction geometrically amplifies actuator stroke without discrete hinges. 

The bellow actuator is positioned at the center of the unit cell and designed for bidirectional actuation under both positive pressure and vacuum. The bellow is joined axially to the struts during parametric modeling, forming a single monolithic elastomeric structure. Mechanical coupling between actuator and lattice occurs through continuous material integration rather than discrete anchoring interfaces.

Contraction of the bellow induces coordinated bending and rotation of the curved struts, distributing deformation across the cell. Due to the anisotropic strut arrangement, struts are laterally interlocked along one diagonal direction within each layer, while remaining free in the orthogonal direction. As individual unit cells are tessellated, this directional interlocking promotes lateral load transfer along one axis while permitting relative freedom along the other, reinforcing the intended axial contraction behavior. 

Each EPUC therefore functions simultaneously as a structural element and an actuation primitive. When tessellated, global deformation emerges from the collective and directionally biased interaction of these unit cells, enabling programmable morphology through selective actuation.

\subsection{Parametric Design and Fabrication}
\label{subsec:parametric}

All geometries were developed within a fully parametric workflow using Grasshopper (Rhino 3D, McNeel, USA), a platform widely adopted for the digital design of architected lattice structures \cite{weegerDigitalDesignNonlinear2019a}. The lattice unit cell was defined through a set of controllable curves describing strut layout and nodal connectivity. These curves were converted into the lattice struts using the \emph{Multipipe} component, enabling continuous control over strut diameter and node blending while preserving smooth transitions throughout the structure.

The pneumatic bellow actuator profile was then defined parametrically and integrated directly into the unit-cell geometry. During tessellation, bellow geometries were boolean-unioned with the lattice struts to form a single continuous body. To enable serial pneumatic connectivity between adjacent cells, lattice curves intersecting the bellow region were locally trimmed at actuator interfaces prior to volumetric conversion, ensuring uninterrupted internal pneumatic pathways across tessellated cells.

Using this framework, $1\times1$, $2\times2$, and $3\times3$ tessellations of the embedded pneumatic unit cell were generated. Geometric parameters including strut curvature, unit-cell size, and bellow dimensions, remained globally adjustable, enabling rapid iteration while preserving actuator-lattice alignment.

All specimens were fabricated using Elastic 50A Resin (Formlabs, USA), a flexible photopolymer suitable for high-strain applications. Samples were printed on a Form 4 stereolithography (SLA) printer (Formlabs, USA) with a layer thickness of 0.1~mm. Exposure parameters were set according to manufacturer recommendations for Elastic 50A resin.

Post-processing followed the standard Formlabs protocol. Printed parts were washed in isopropyl alcohol for 20~min using a Form Wash station, followed by UV curing at 70$^\circ$C for 30~min in a Form Cure unit to ensure complete polymerization.

\begin{figure*}[t]
    \centering
    \vspace{1em}
    \includegraphics[width=0.9\textwidth]{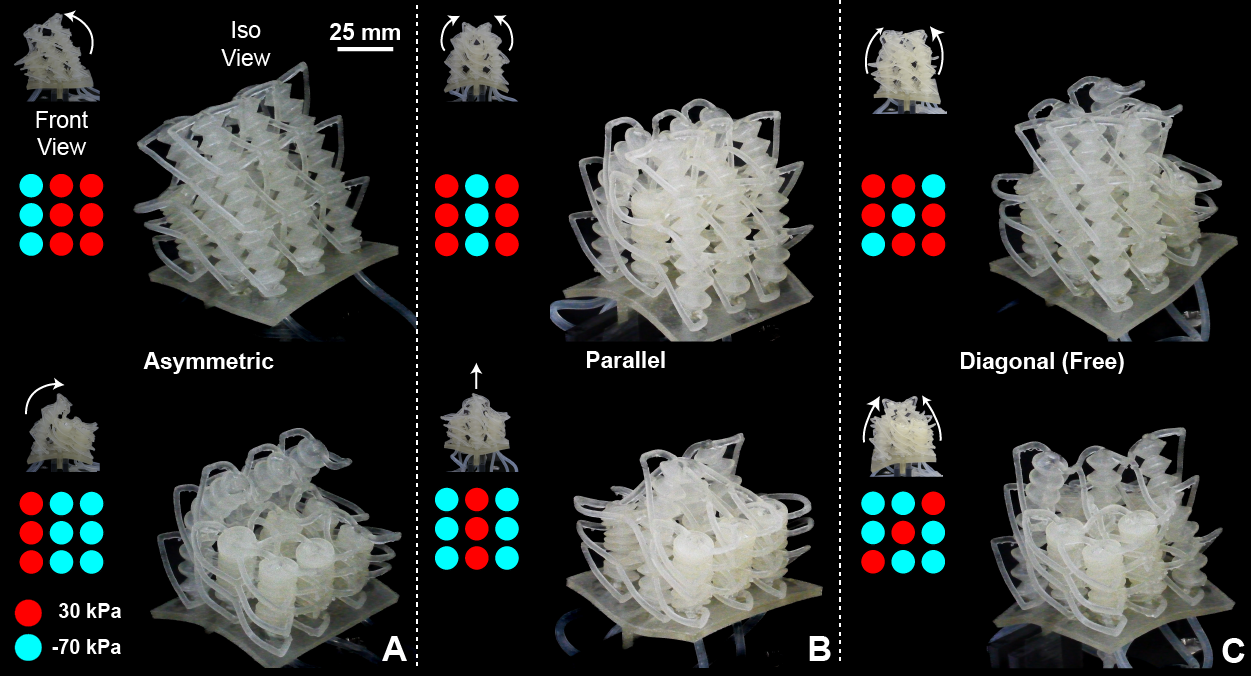}
    \caption{Morphological variation via selective unit-cell actuation in a $3\times3\times3$ lattice. Colored grids indicate commanded pressures for each vertical column of embedded pneumatic unit cells (red: +30~kPa, cyan: -70~kPa). Symmetric actuation about horizontal, vertical, or diagonal centerlines produces convergent contraction suitable for grasping-like configurations. Asymmetric actuation generates differential shortening between columns, resulting in global bending. The anisotropic curved-strut geometry leads to distinct deformation modes depending on actuation topology.}
    \label{fig:motionmodes}
\end{figure*}

\subsection{Experimental Protocol}
\label{subsec:exp-protocol}

To evaluate actuation performance and scalability, specimens comprising $1\times1$, $2\times2$, and $3\times3$ tessellations of the EPUC were characterized using a universal testing machine (ZwickRoell, Germany). For each sample, top and bottom interface plates were integrated to enable clamping.

Pressure was regulated via pressure and vacuum regulators and commanded through a custom Python interface using a data acquisition board (NI USB-6218, National Instruments, USA). For the $1\times1$, $2\times2$, and $3\times3$ tessellations used in UTM tests, all embedded pneumatic unit cells were connected to a single pneumatic manifold (one pressure line), such that all cells were driven simultaneously. A solenoid valve was used to switch between positive pressure and vacuum sources. A sinusoidal pressure waveform spanning $-70$~kPa to $+30$~kPa at 0.2~Hz was applied in all UTM experiments. Pressure signals and the corresponding displacement and force measurements from the UTM were synchronized and recorded for analysis.

For selective-actuation experiments, the nine columns of the $3\times3\times3$ lattice were pneumatically grouped according to the desired activation topology. Two pressure lines (+30~kPa and -70~kPa) were supplied through solenoid valves that switched airflow between positive pressure and vacuum. Each column was connected to one of the two lines during a given trial, enabling spatially patterned actuation.

Three testing modalities were performed. In isotonic tests, a constant preload of 1~N was applied while allowing axial motion, and displacement was recorded under the commanded pressure waveform. In isometric tests, axial position was held constant while the resulting force response was measured. Finally, free-moving cyclic tests were conducted with 0~N preload for 500 pressure cycles to evaluate repeatability and durability, with displacement recorded throughout.

\section{EXPERIMENTAL RESULTS}

\subsection{Embedded Unit-Cell Actuation Performance}
\label{subsec:performance}

Fig.~\ref{fig:exp-setup}B shows the isotonic displacement response of $1\times1$, $2\times2$, and $3\times3$ tessellations under a sinusoidal pressure input spanning $-70$~kPa to $+30$~kPa with a constant 1~N preload. All configurations exhibit repeatable bidirectional actuation, with contraction under vacuum and extension under positive pressure. Peak positive-pressure displacement increased from 2.6~mm ($1\times1$) to 3.4~mm ($2\times2$) and 4.8~mm ($3\times3$). Peak vacuum contraction increased from $-7.3$~mm to $-16.7$~mm and $-26.2$~mm, respectively, indicating that stroke scales with the number of serially connected unit cells. Hysteresis was more pronounced during vacuum-driven contraction; as the lattice compacted, strut contact increased the effective stiffness, producing a nonlinear pressure--displacement response.

Isometric force measurements (Fig.~\ref{fig:exp-setup}C) show the same scaling with tessellation size. Under vacuum, peak contraction force increased in magnitude from $-2.2$~N ($1\times1$) to $-7.4$~N ($2\times2$) and $-15.9$~N ($3\times3$). Under positive pressure, peak expansion force increased from 0.9~N to 2.7~N and 4.6~N, respectively. The larger forces observed during vacuum-driven contraction are consistent with the contracted configuration becoming stiffer as struts engage.

Free-moving cyclic tests over 500 pressure cycles (Fig.~\ref{fig:exp-setup}D) showed stable displacement amplitude after a short initial transient. No visible structural failure was observed over the test duration, indicating repeatable actuation without measurable fatigue-induced degradation in this time window.

Overall, both displacement and force increased with tessellation size. Vacuum actuation produced larger stroke and higher force, along with stronger hysteresis, due to lattice compaction and strut contact during contraction.

\begin{figure*}[t]
    \centering
    \vspace{1em}
    \includegraphics[width=0.9\textwidth]{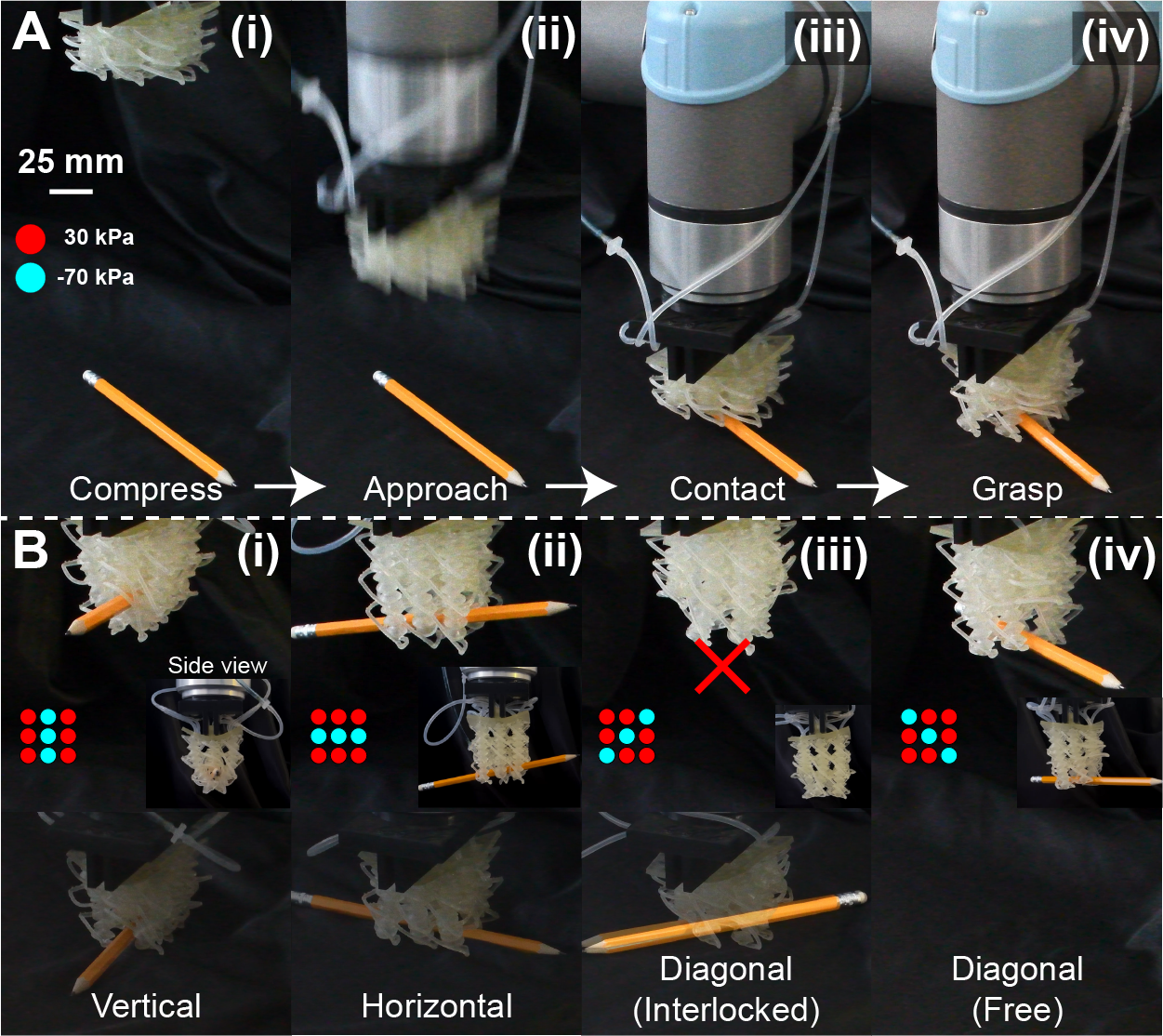}
    \caption{Selective actuation enables directional grasping through distinct actuation topologies for the $3\times3\times3$ array. Colored grids indicate commanded pressures for each unit cell (red: +30~kPa, cyan: -70~kPa). Semi-transparent overlays illustrate the deformation envelope during grasp. (A) Temporal sequence for the diagonal (free) configuration: (i) compress, (ii) approach, (iii) contact, and (iv) grasp. (B) Grasp outcomes under different actuation topologies: (i) vertical, (ii) horizontal, and (iv) diagonal (free) configurations result in successful object retention, whereas (iii) diagonal (interlocked) leads to failed object capture due to geometric interlocking.}
    \label{fig:grasps}
\end{figure*}


\subsection{Morphological Variation via Selective Actuation}

Morphological behavior was evaluated using a $3\times3\times3$ tessellation of the embedded pneumatic unit cell (Fig.~\ref{fig:motionmodes}). The array was fixed at the base while all upper surfaces remained unconstrained. Individual columns were independently actuated with either +30~kPa or -70~kPa.

Asymmetric actuation produced global bending. When one column was contracted under vacuum while adjacent columns were extended, differential axial shortening generated curvature toward the actuated side. The resulting deformation was smooth and continuous, with bending localized primarily along the actuated column.

Parallel (symmetric) actuation about horizontal or vertical centerlines produced axis-aligned inward contraction. Opposing columns shortened equally, generating a centered convergent configuration resembling a gripper-like closure.

Symmetric diagonal actuation also produced inward contraction, but with a rotated deformation axis. Although the pressure pattern is symmetric, the response reflects the anisotropic strut arrangement described in Section II-A. Within each layer, struts are laterally interlocked along one diagonal direction while remaining free along the orthogonal direction. Actuation along the free diagonal permits relative sliding between columns and results in compliant inward contraction without lateral constraint.

These observations indicate that deformation mode depends on both actuation symmetry and the directional mechanical coupling introduced by the tessellated unit-cell geometry.

\begin{figure*}[t]
    \centering
    \vspace{1em}
    \includegraphics[width=0.9\textwidth]{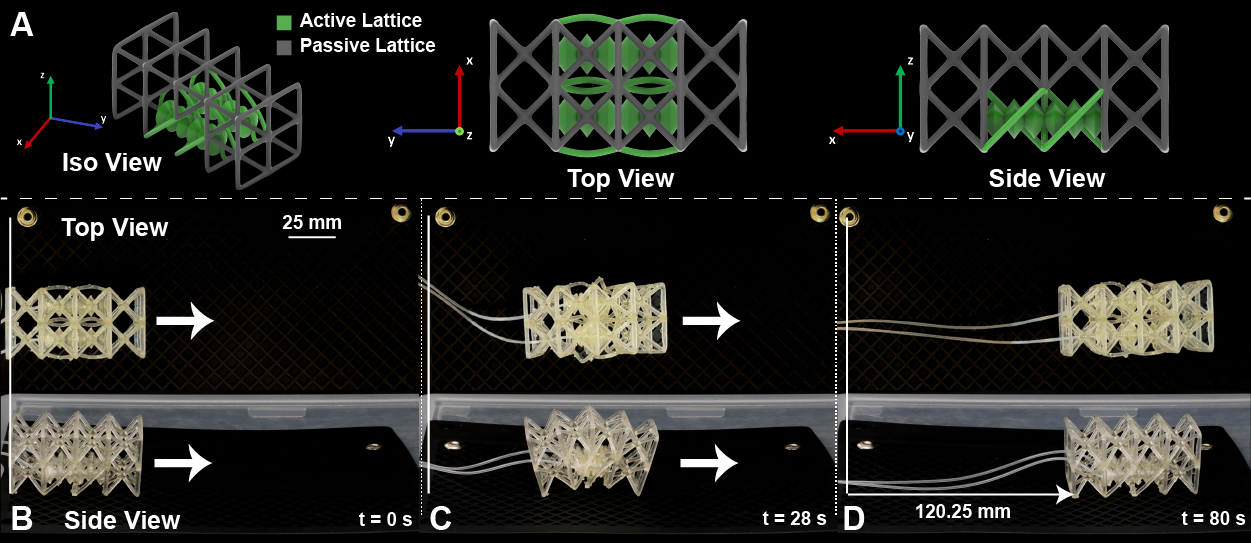}
    \caption{Crawler of a $1\times2\times2$ embedded pneumatic unit-cell tessellation operating in a bending mode and supported by a passive edge octahedral lattice. (A) 3D render of the crawler with various views. Top and side views illustrate progression from the initial configuration (B, $t=0$ s) to intermediate bending (C, $t=28$ s) and final translated state (D, $t=80$ s).}
    \label{fig:locomotion}
\end{figure*}

\section{Functional Demonstrations}


\subsection{Directional Grasping}

The $3\times3\times3$ lattice was mounted to a UR5e robotic arm (Universal Robots, Denmark) and used to grasp a cylindrical object (Fig.~\ref{fig:grasps}). Individual columns were independently commanded with either vacuum or positive pressure.

Symmetric vertical and horizontal actuation patterns produced successful grasping along the corresponding axes. Balanced contraction generated inward curvature that enclosed and retained the object.

Two diagonal actuation patterns were also evaluated. Owing to the anisotropic lateral connectivity of the curved struts, interlocking occurs preferentially along one diagonal direction within each tessellated layer. In the free diagonal configuration, relative sliding between adjacent columns was permitted, producing inward contraction and successful object retention. In the interlocked diagonal configuration, lateral contact between neighboring struts restricted transverse motion. This constraint altered the deformation pathway and prevented effective enclosure of the object, resulting in grasp failure (Fig.~\ref{fig:grasps}B(iii)).

These results show that grasp performance depends not only on actuation symmetry but also on the directional mechanical constraints introduced by unit-cell interlocking.

\subsection{Bending-Driven Locomotion}
To demonstrate locomotion, the EPUCs were integrated with passive unit cells to form a hybrid architecture (Fig.~\ref{fig:locomotion}A). The passive cells were selected based on the Edge Octa topology, which exhibits low axial tension and favorable bending compliance in lattice-based structures \cite{schouten3DPrintableGradient2025b}. The final configuration consisted of a $2\times2\times4$ tessellation, with a central $1\times2\times2$ tesselation of EPUCs oriented downward and surrounded by Edge Octa unit cells. This localized actuation within the core of the structure while the surrounding passive cells provided compliant support and allowed for global bending.

The structure was equipped with an asymmetric foot and placed on a patterned substrate (Fig.~\ref{fig:locomotion}B). This asymmetry enabled forward displacement during contraction while limiting backward slip during relaxation (Fig.~\ref{fig:locomotion}C). 

The crawler was connected to the same pneumatic system described previously and driven using a sinusoidal pressure signal ranging from $-100$~kPa to $0$~kPa at a frequency of 0.3~Hz. This cyclic bending-release sequence generated a crawling gait, allowing the robot to crawl 120.25~mm in 80~s (Fig.~\ref{fig:locomotion}D). These results demonstrate that selective unit-cell actuation, when combined with passive lattice elements, enables functional locomotion within a monolithic architected structure.

\section{DISCUSSION}

The presented lattice architecture replaces modular segmentation with distributed actuation at the unit-cell level. Conventional continuum soft robots often rely on rigid plates or discrete interfaces between actuation segments, introducing stiffness discontinuities and constraining curvature to predefined locations. In contrast, the embedded pneumatic unit cell integrates structure and actuation within a single repeatable element. When tessellated, curvature and force generation emerge from coordinated cellular actuation rather than from mechanical boundaries. This suggests a shift from segment-based kinematics toward continuously actuated architectures in which spatial actuation patterns shape deformation gradients.

Beyond uniform tessellation, the proposed architecture naturally enables spatial variation of unit-cell geometry across the lattice. Because the structure is defined parametrically, gradients in strut diameter, curvature, node connectivity, or unit-cell size can be introduced to create heterogeneous stiffness and anisotropy within a single monolithic body. Such spatial grading has been shown to enable musculoskeletal-inspired behavior and multi-stiffness robotic structures in lattice-based systems \cite{guanLatticeStructureMusculoskeletal2025,schouten3DPrintableGradient2025b}. Within the present framework, graded geometry could bias curvature toward predefined regions, localize bending, or redistribute stress during actuation without modifying the global actuation pattern. Coupling functional gradients with selective unit-cell actuation offers a pathway toward architected soft robots in which both compliance and deformation are spatially encoded through structural design. The locomotion demonstration further illustrates how heterogeneous tessellation, combining active and passive unit cells, can introduce functional asymmetry at the architectural level.

The curved-strut unit cell also highlights anisotropy as an explicit design variable. Because deformation is bending-dominated and orientation-dependent, actuation along different axes produces distinct global responses. This directional behavior arises from the asymmetric lateral connectivity of the struts, which promotes load transfer along one diagonal while maintaining compliance along the orthogonal direction, as a result, anisotropy is a controllable feature that can be leveraged through actuation topology. Future unit-cell designs could intentionally incorporate multistable elements, snap-through mechanisms, or origami- and kirigami-inspired folding geometries \cite{rahmanReprogrammableSoftPneumatic2025,liangBifurcationBistabilityPneumatically2023}. Embedding instability-driven transitions within the unit-cell geometry may enable discrete state switching, stiffness modulation, or mechanically encoded memory within the lattice itself.

Scaling this approach introduces challenges in pneumatic routing and addressability. Increasing spatial resolution improves morphological fidelity but requires additional control channels and manifold complexity. Hierarchical fluidic routing, multiplexed pressure control, or embedded microfluidic networks may be required for large-scale implementations. Addressability therefore becomes a system-level design problem alongside geometric co-design.

Finally, the parametric framework developed here is naturally extensible to embedded sensing. Optical waveguides, strain-sensitive paths, or pressure-based self-sensing chambers could be incorporated directly into the unit-cell geometry \cite{truninDesign3DPrinting2025,trubyFluidicInnervationSensorizes2022b}. Integrating sensing at the same architectural scale as actuation would enable closed-loop morphological control, where deformation state informs actuation topology in real time. Such integration would further dissolve the boundary between structure, actuation, and sensing in lattice-based soft robotic systems.

Taken together, these results suggest that lattice robots are better described as architected materials whose global behavior emerges from spatially distributed cellular actuation.

\section{CONCLUSION}

This work presented a pneumatic lattice architecture in which actuation and structural geometry are co-designed at the unit-cell level. The embedded pneumatic unit cell integrates curved-strut compliance and a bidirectional bellow within a single monolithic element, eliminating discrete actuator-lattice interfaces. When tessellated, global deformation emerges from coordinated cellular interaction rather than from modular segmentation.

Experimental characterization demonstrated scalable displacement and force generation across $1\times1$, $2\times2$, and $3\times3$ configurations, with repeatable performance under cyclic loading. Selective actuation of unit cells enabled distinct global morphologies, including bending and grasping, without modifying the physical structure.

These results demonstrate that actuation topology directly governs global deformation in lattice-based soft robots. By embedding actuation at the unit-cell scale, structure and actuation become intrinsically coupled within a single repeatable element. When tessellated, global behavior emerges from coordinated cellular interactions rather than from modular segmentation or discrete interfaces. This framework establishes a scalable approach to spatially distributed actuation in monolithic soft robotic structures.

\addtolength{\textheight}{-16cm}   






\bibliographystyle{IEEEtran}

\bibliography{biblio}

\end{document}